\documentclass[conference,letterpaper]{IEEEtran}
\IEEEoverridecommandlockouts

\usepackage[style=ieee,backend=biber]{biblatex}
\def\IEEEtitletopspace{18pt}

\usepackage{amsmath,amssymb,amsfonts}
\usepackage{graphicx}
\usepackage{textcomp}
\usepackage{xcolor}
\usepackage{url}
\usepackage{todonotes}
\usepackage{hyperref}

\begin{document}

\title{Characterizing Refraction-Induced Ranging Bias in 
Underwater Collaborative Localization}

\author{
\IEEEauthorblockN{Timothy Kogucki}
\IEEEauthorblockA{
\textit{KTH Royal Institute of Technology}\\
Stockholm, Sweden\\
tkogucki4@gmail.com}
\and
\IEEEauthorblockN{Alan Papalia}
\IEEEauthorblockA{
\textit{University of Michigan}\\
Ann Arbor, MI, USA \\
apapalia@umich.edu}
}

\maketitle
\vspace{-30mm}

\begin{abstract}

This work studies how refraction-induced bias on acoustic ranging affects
multi-agent collaborative localization in a range of oceanographic conditions
and spatial scales.
While multi-agent range-aided navigation, which uses range measurements to
either fixed infrastructure or other agents, is a promising solution to the
challenges of large-scale underwater localization, its accuracy depends strongly
on the quality of range measurements.
Sound speed variability induces refraction (bending) of acoustic rays, yet, for
algorithmic tractability, standard sensor fusion pipelines assume straight-line
propagation.
This refraction systematically biases range measurements to be longer than the
straight-line assumption predicts.
However, the effects of this bias on multi-agent collaborative localization on
kilometer scales remains unexplored.
We present a series of simulated experiments with several agents operating over
kilometer scales. The simulation uses HYCOM reanalysis data to recreate
realistic oceanographic conditions, ray tracing to generate refraction-informed
ranges, and a centralized multi-agent factor graph estimator to quantify the
resulting measurement bias on estimated trajectories.
Preliminary results indicate that refraction-induced bias can induce
significant degradation of estimated trajectories, particularly in regions with
sharp sound-speed gradients. We also share the simulation environment to
support further studies \url{https://github.com/UMich-RobotExploration/manta-ray}.

\end{abstract}

 
\section{Introduction}

Modern ocean exploration increasingly relies on distributed autonomous
platforms capable of capturing dynamic oceanographic processes across
large spatial and temporal scales. Autonomous underwater platforms
enable dense sampling without continuous vessel support, but
multi-agent operations depend on precise navigation in
global navigation satellite system (GNSS) denied environments where inertial
drift and feature sparsity
impose localization constraints \cite{PaullIJOE2014}.

Range-aided simultaneous localization and mapping (RA-SLAM) addresses these
challenges by fusing inertial, pressure, velocity, GNSS, and acoustic
measurements to estimate vehicle pose \cite{papalia2024certifiably}.
Range sensing has a distinct advantage in localization systems: since measurement
data-associations are typically known, these measurements avoid the common
brittleness due to ambiguous data-association found in other sensing types.
\cite{rosen_advances_2021}. 
Additionally, RA-SLAM allows using other agents or deployed beacons to augment
or replace the need for detectable features in the environment \cite{papalia2022prioritized}.
However, RA-SLAM in underwater environments poses a unique challenge: 
the ocean's sound-speed varies due to differing distributions of temperature,
salinity, and pressure \cite{JensenBOOK2011}. This causes refraction of acoustic
ranging.
Propagation is typically simplified in
localization systems to straight lines (a uniform sound-speed
assumption) \cite{PaullIJOE2014,BahrIJRR2009}. When this assumption
is violated, the range component of the fused estimate is
\emph{systematically} biased rather than merely noisy -- i.e., the
error is not zero-mean. 
Sound speed profile (SSP) informed
localization has been studied for long-baseline systems that use
fixed beacons with known locations
\cite{mikhalevsky_deep_2020,van_uffelen_localization_2016}, but
physics-informed ranging remains underexplored for modern
multi-agent factor-graph pipelines operating at large scales. 

This
work quantifies how the systematic bias introduced by the
straight-line assumption affects the estimated trajectories of a
multi-agent team.
As our primary focus is the effect of biased measurements on the estimator,
we perform state estimation in a centralized, offline fashion.
To generate scenarios of interest we developed an 
open-source simulator, \texttt{MantaRay}, capable of ingesting HYCOM reanalysis products \cite{HYCOM} and simulating acoustic refraction between a fleet of agents.

\begin{figure}[t]
    \centering
    \includegraphics[width=0.97\linewidth]{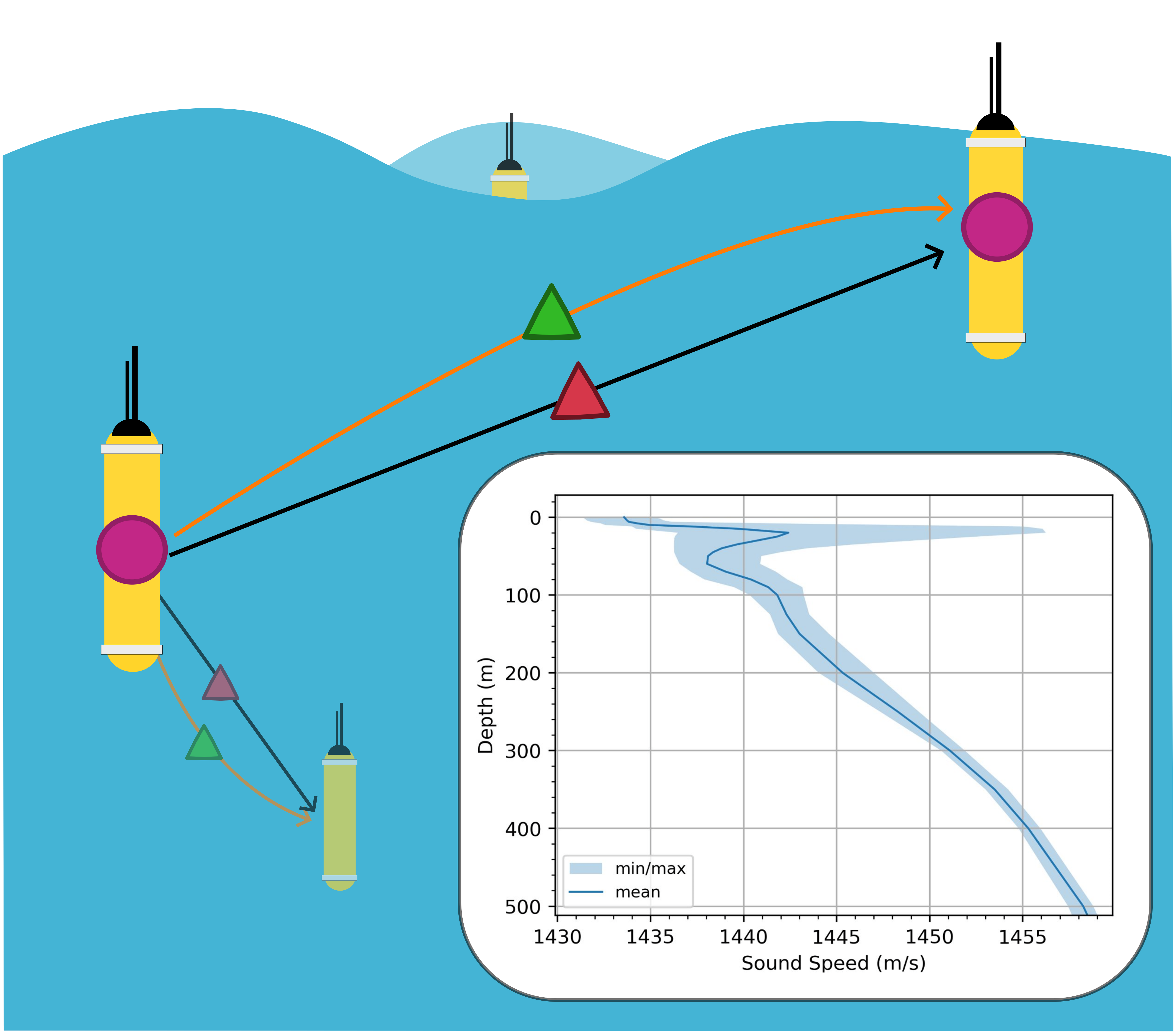}
    \caption{Sound speed varies with depth, longitude, and latitude,
    causing acoustic waves to follow curved refracted paths (orange)
    rather than the straight-line paths (black) assumed by typical
    range-aided localization systems.}
    \label{fig:refraction}
    \vspace{-5mm}
\end{figure}

\begin{figure}
    \centering
    \includegraphics[width=0.99\linewidth]{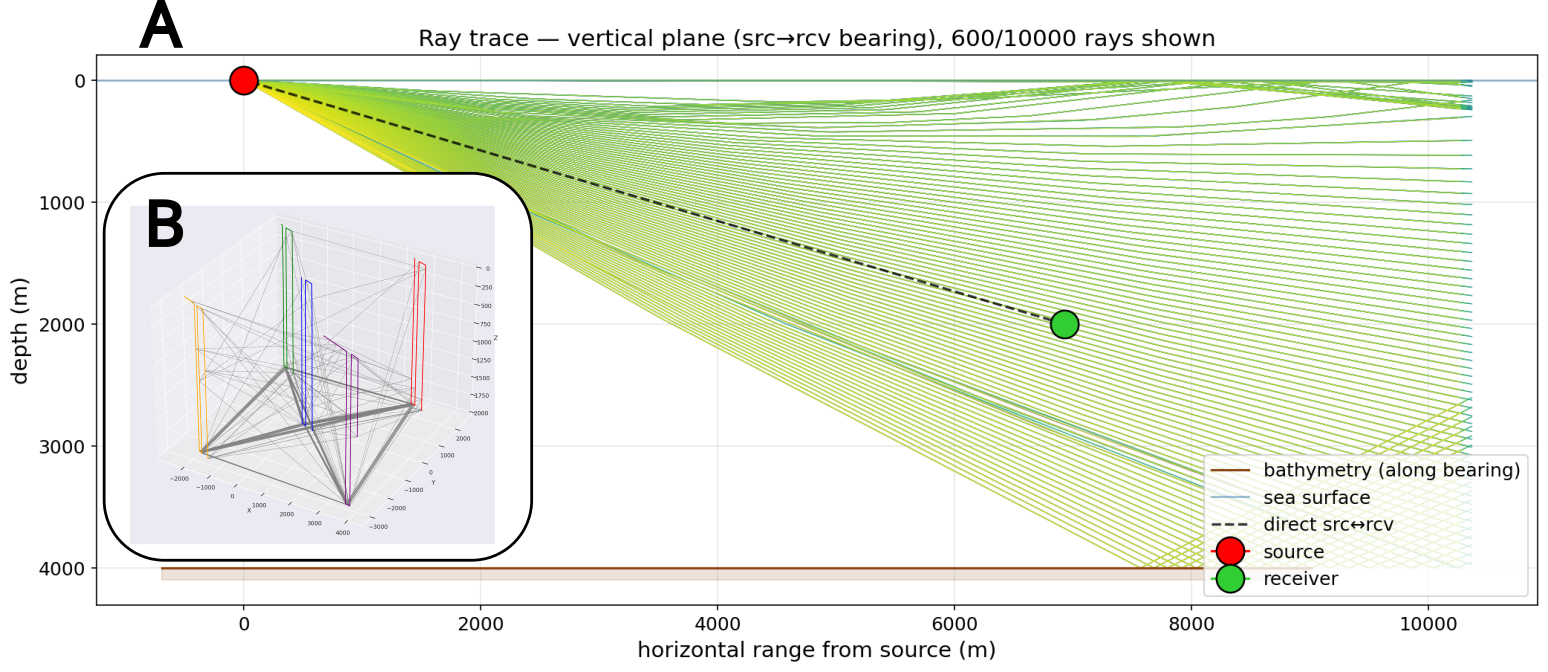}
    \caption{\texttt{MantaRay} simulation artifacts for a Beaufort Sea
    scenario. (A) Ray-traced acoustic propagation for a single
    multi-agent pairing, showing refraction through the water column,
    a near-surface sound-speed duct, and reflections from bathymetry.
    (B) The resulting multi-agent factor graph: floats (colored) are
    connected by an inter-agent range network (grey) spanning pairings
    at both matched and disparate depths.
    }
    \label{fig:results}
    \vspace{-3mm}
\end{figure}

We summarize the contributions of this work as follows:
\begin{itemize}
\item A characterization of refraction-induced ranging bias in underwater multi-agent localization.
\item An open source \texttt{MantaRay} simulator capable of capturing how oceanographic properties effect acoustic ranging in a fleet of agents.  
\end{itemize}

\section{Related Work}
We briefly review works related to range-based underwater localization, SSP-informed localization, marine robotics simulation, and modeling acoustic propagation.

\subsection{Range-Based Underwater Localization}
Range-based localization systems historically have leveraged external acoustic infrastructure. Such acoustic infrastructure is commonly classified by the spacing of their receiver arrays: Ultra Short Baseline (USBL), Short Baseline (SBL), and Long Baseline (LBL) \cite{paull_auv_2014}. 
USBL and SBL systems are often deployed on vehicles, while LBL systems are typically viewed as fixed infrastructure that fences in agents operating range, though this is not necessarily required. 
Modern collaborative approaches leverage inter-agent ranging and information exchange to reduce the need for fixed infrastructure \cite{paull_auv_2014, paull_communication-constrained_2015}. Collaborative approaches fall into two broad categories: centralized and decentralized. While more robust to communication failures, decentralized methods introduce challenges around consistency, data association, and double-counting of
shared information \cite{bahr_cooperative_2009}.
We will not discuss decentralized approaches (e.g.:
\cite{velasco_factor-graph-based_2025,
arrichiello_opportunistic_2012,bahr_cooperative_2009, fallon_measurement_2010,
paull_decentralized_2014}) as the implemented framework in this work is
centralized, but these works address important distributed needs of multi-agent
systems. Centralized multi-agent systems have been explored by
\cite{paull_communication-constrained_2015} with \cite{paull_decentralized_2014}
highlighting the major challenges: low-bandwidth communication and communication
dropouts.  

\subsection{SSP-Informed Underwater Localization and Simulation}
Refraction-aware acoustic navigation has largely developed in the
context of fixed long-baseline (LBL) infrastructure and basin-scale
ocean acoustic tomography, where well-characterized SSPs and broadband signals support acoustic propagation modeling 
against a known environment or joint inversion of the SSP and
receiver position \cite{mikhalevsky_deep_2020, duda_evaluation_2006,
van_uffelen_localization_2016}. These studies demonstrate the value
of accounting for refraction, but the enabling setups (dedicated
arrays, broadband sources, offline processing) differ meaningfully
from the on-vehicle regime targeted by a mobile multi-agent fleet.
Less attention has been paid to how the straight-line assumption
propagates into errors of the estimated trajectories of a modern multi-agent
factor-graph pipelines in oceanographic conditions, which is the gap this work addresses.

\subsection{Marine Robotics Simulation}
One of the challenges of this work is that no existing marine robotics simulator supports the study of refraction effects at a distributed multi-agent scale.
Existing simulators \cite{CieslakOCEANS2019,PotokarICRA2022,kartasev_smarcsim_2025} provide valuable capabilities such as AUV hydrodynamics and sonar simulation, but are not designed to represent spatially-varying oceanographic properties or their effect on acoustic sensing.
This work presents the first iteration to address this capability gap. 

\subsection{Acoustic Propagation Tools}
Ocean acoustic propagation is governed by well-established physics,
and several families of numerical solvers have been developed to
approximate it: spectral (FFP) \cite{JensenBOOK2011}, normal-mode \cite{porter_kraken_1992}, ray \cite{porter_bellhop_2011}, parabolic-equation \cite{tappert2005parabolic},
and finite-difference methods \cite{JensenBOOK2011}. These families differ chiefly in the
frequency regime they target and in whether they treat the environment
as range-independent or range-varying \cite{JensenBOOK2011}. Ray
methods are typically the practical choice above roughly a kilohertz \cite{JensenBOOK2011}
the regime where most vehicle mounted acoustic sensors operate in (e.g. \cite{gallimore_whoi_2010}).
These methods also allow variation of oceanic properties in all three dimensions, making them well suited for the work presented here \cite{JensenBOOK2011}.
Bellhop \cite{porter_bellhop_2011} is a widely used open-source implementation
of a ray method solver.
Recent efforts such as
\texttt{bellhopcuda} \cite{PishaJASA2023} implement ray-method
solvers in C++/CUDA with GPU acceleration, allowing the runtime requirements to embed ray-traced propagation inside larger simulation
pipelines.

\section{Methodology}
The simulation environment we develop,
\texttt{MantaRay}, comprises of three principal components: oceanographic data
processing, acoustic propagation via ray tracing, and prescribed-velocity kinematics model.
These three components generate measurements and ground truth poses, which are
then corrupted \emph{post hoc} with various noise models and constructed into a centralized factor graph for estimation.
The effects of refraction are evaluated by conducting a Monte Carlo study where 25 repeated trials of the same mission are conducted with independently sampled noise realizations. 

\begin{figure*}[t]
    \centering
    \includegraphics[width=0.85\textwidth]{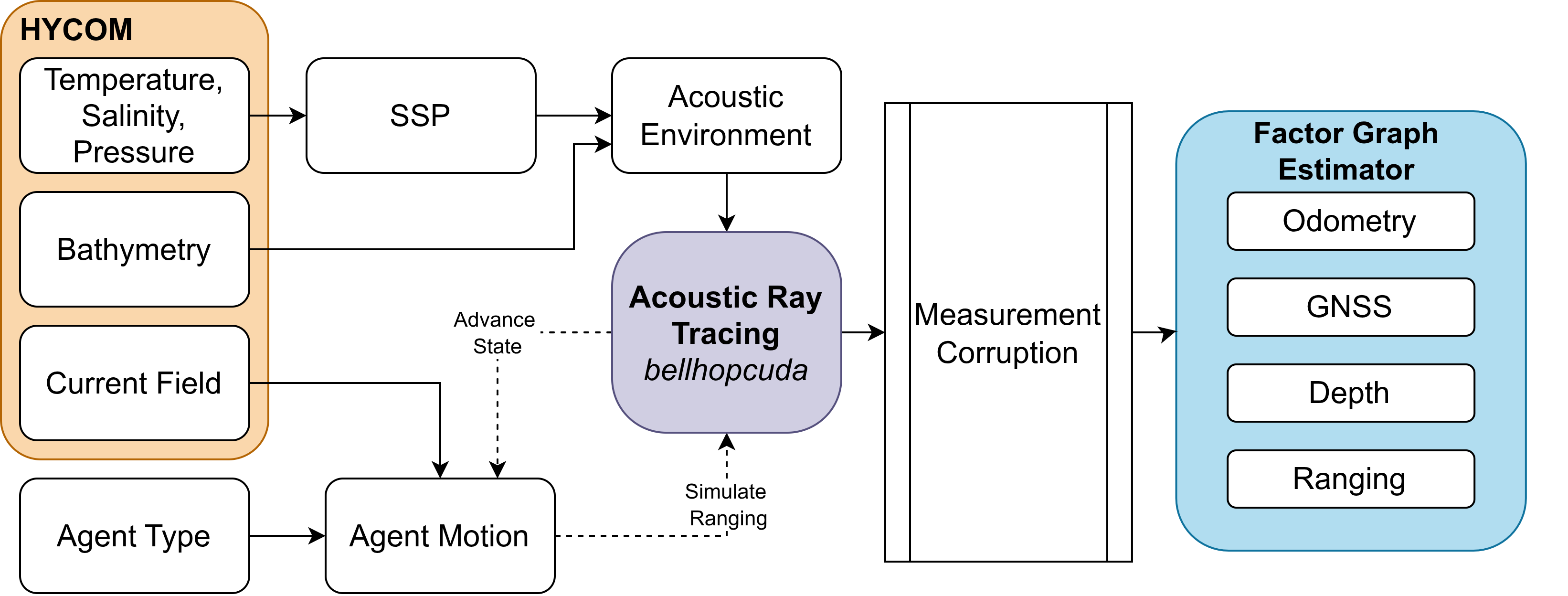}
    \caption{\texttt{MantaRay} utilizes HYCOM re-analysis ocean environment data. Oceanic data is utilized to compute inter-agent ranging for a collaborative fleet of agents such that the effects of acoustic refraction can be evaluated in a centralized factor graph estimator.}
    \label{fig:method}
\end{figure*}
  
\subsection{Acoustic and Rigid Body Simulation}
Oceanographic data (temperature, salinity, pressure, and
currents) is sourced from HYCOM reanalysis products and is used to construct
three-dimensional sound-speed grids and coarse bathymetry. 
Course bathymetry is utilized to enable reflections from the sea floor while the surface of the water is modeled as a plane with no surface undulations.
\texttt{MantaRay}'s oceanographic pipeline is currently capable of including any region covered by HYCOM databases or data that can be converted to match the application programming interface (API).
HYCOM data does not natively provide sound speed, requiring post-processing that estimates it from temperature, salinity, and pressure \cite{McDougallSCOR2011}.
Another thing to note is that HYCOM fidelity is limited in under-ice Arctic regimes; sustained
under-ice studies would benefit from ingesting other data, e.g. ice tethered profiler (ITP) data.

Acoustic propagation is simulated with a GPU-accelerated ray tracer, \texttt{bellhopcuda}, to compute
time-of-flight along both direct and multi-path arrivals \cite{PishaJASA2023}. 
Although \texttt{MantaRay} is capable of simulating multi-path arrivals, all simulated experiments in this work only used direct range measurements.
To obtain time-of-flight range measurements, a single \texttt{bellhopcuda} environment is constructed for each multi-agent fleet where aforementioned SSP, bathymetry, and ocean surface grids are encoded from HYCOM data.
Throughout the simulation only agent positions and beam numbers are altered: the former to enable evaluation of all range pairings and the latter to enable iterative computation of fastest acoustic arrivals.
Each acoustic propagation step is run with static agent positions with the assumption that agent motion during the ranging period is an insignificant distance compared to overall ranging.
Discretization of acoustic rays is parameterized via a ray cone and beam number. The ray cone was set to a $15^\circ$ symmetrically along the vector between source and receiver agents to balance beam density while ensuring there was sufficient spread to hit the receiver. 
Beam count is iteratively incremented to aid in finding direct-path arrivals or to evaluate for converging multi-path arrivals.
Acoustic simulations were done with no absorption modeling and lossless reflections from the sea floor and water surface.
Due to the structure of \texttt{bellhopcuda}'s solver, it was preferred to simulate reflections and simply filter them out of arrival data then attempt to create a reflection free environment.
All presented results allow only direct-path arrivals with no reflections to evaluate systematic bias from refraction.
Evaluation of acoustic multi-pathing effects on factor graph estimators is deferred to future work.
As \texttt{MantaRay} interfaces with a fully featured \texttt{bellhopcuda} library, more complex phenomena can be modeled with Bellhop in the future (e.g., seafloor models and water volume absorptivity).

Rigid-body kinematics are simulated in full 6-DOF to enable future extensions. 
Simulated platforms take twist ($\nu= [\omega, v] \in \mathbb{R}^6$) as their input, which the solver integrates to compute both position and orientation.  
The platforms used in this study are constrained to 3-DOF by only applying translational velocities as inputs to evaluate positional ($x,y,z$) estimates only. 

The simulated platforms are homogeneous float-like devices with
odometry, depth sensing, intermittent surface GNSS, and acoustic
ranging modems, operated as a team under a centralized estimator. These platforms aim to mimic Argo floats \cite{roemmich2009argo} in their controls and passive drift while providing additional sensing modalities to enable underwater communications and sensing. Within \texttt{MantaRay}, passive drifting is achieved by directly interpolating current fields and applying them as translational inputs, while depth is maintained with a simple state machine and position based proportional controller that has a saturation limit set at the maximum descent velocity prescribed.

\subsection{Estimator and Measurement Perturbation}
The centralized estimator is built on a single
factor graph combining all of the agents and implemented in GTSAM \cite{DellaertGH2022}.
Factor graphs represent the joint posterior over agent poses as a product of
local factors (i.e., measurements). This naturally accommodates asynchronous
measurements both onboard and across cooperating vehicles.

The estimator is evaluated with two ways of generating the range measurements. In the baseline, synthetic
ranges are drawn from Euclidean inter-agent distances,
\begin{equation}
r_{ij} = \left\|x_i - x_j\right\| + \eta, \eta \sim \mathcal{N}(0, \sigma^2)
\end{equation}
matching the estimator's internal model. In the refraction-informed
condition, ranges are computed from ray-traced travel time
$t_{\text{ToF}}$ scaled by the sound speed sampled at the transmitting
agent's position,
\begin{equation}
r_{ij} = c_{\text{agent}} \cdot t_{\text{ToF}}~+~\eta.
\end{equation}

The ranging measurement model used in the simulator, follows standard practice
(to enable computational tractability) and uses the same straight-line range
factor in both cases. 
Because the estimator is unaware of refraction in either case, any difference
between estimates is due to the straight-line-propagation assumption being
invalidated when refraction occurs.
The backend assembles one centralized factor graph over all agents and solves it in a single batch in GTSAM \cite{DellaertGH2022}.
Variables are a per-agent pose chain in $\mathrm{SE}(3)$ along with a static position for each acoustic beacon when landmarks are present. A factor graph is constructed by parsing \texttt{MantaRay}'s text-based factor graph representation in Python. 
Factors are divided into: pose priors on each agent's initial pose and on
every GNSS surface fix, priors on the landmarks, between factors carrying
recorded odometry, a range factor per acoustic observation, and a
depth prior per pose. The depth prior is deliberately anisotropic, tight on the
world-frame vertical coordinate and effectively unconstrained
horizontally because accurate depth sensing is readily available but provides no information on horizontal positioning. 
All variables are initialized at ground truth so that
measurement error, rather than initialization quality, is the
controlled variable.

The simulator emits noise-free measurements, so a separate stage applies
sensor noise before the graph is assembled. Ranges, depths, and
GNSS
fixes each receive additive zero-mean Gaussian noise; for ranges this
lumps
modem timing jitter into a single $\sigma_{r}$. Odometry is perturbed on the manifold
instead,
with a Gaussian noise in the tangent-space $\xi \sim \mathcal{N}(0,
\mathrm{diag}(\sigma_i^{2}))$
right-composed onto each ground-truth relative pose and per-component
standard deviation

\begin{equation}
  \sigma_{i} = \sqrt{(f_{i}\, \lVert m_{i} \rVert)^{2} + (d_{i}\, \Delta
t)^{2}},
  \label{eq:composite_noise}
\end{equation}

\noindent combining a velocity-scale fraction $f_{i}$ on the per-edge motion
magnitude
$\lVert m_{i} \rVert$ with a drift rate $d_{i}$ scaled by the inter-pose
interval $\Delta t$. 
The unique noise modeling approach was implemented because float agents spend long intervals at near-zero velocity, and a pure velocity-scale model would drive the between factor to be excessively tight.
In order to understand the system-level effects, the scenario is repeated across $25$
realizations.
Within each realization the graph is solved twice, once per measurement
condition. Both conditions are solved under identical odometry, depth, GNSS, and range-noise draws,
so the two solves differ only in the deterministic component of the range.
Guaranteeing each solve receives identical noise draws ensures that any differences shown can be attributed to refraction. 
Trajectory accuracy is reported as absolute translation error (ATE)\cite{sturm_benchmark_2012} against ground truth
, computed via \texttt{evo} \cite{grupp_evo_2017}.
For aggregate ATE distributions, all agents and realizations are combined in to a single distribution for each measurement condition.

\section{Simulated Experiments}
\begin{figure}
    \centering
    \includegraphics[width=0.97\linewidth]{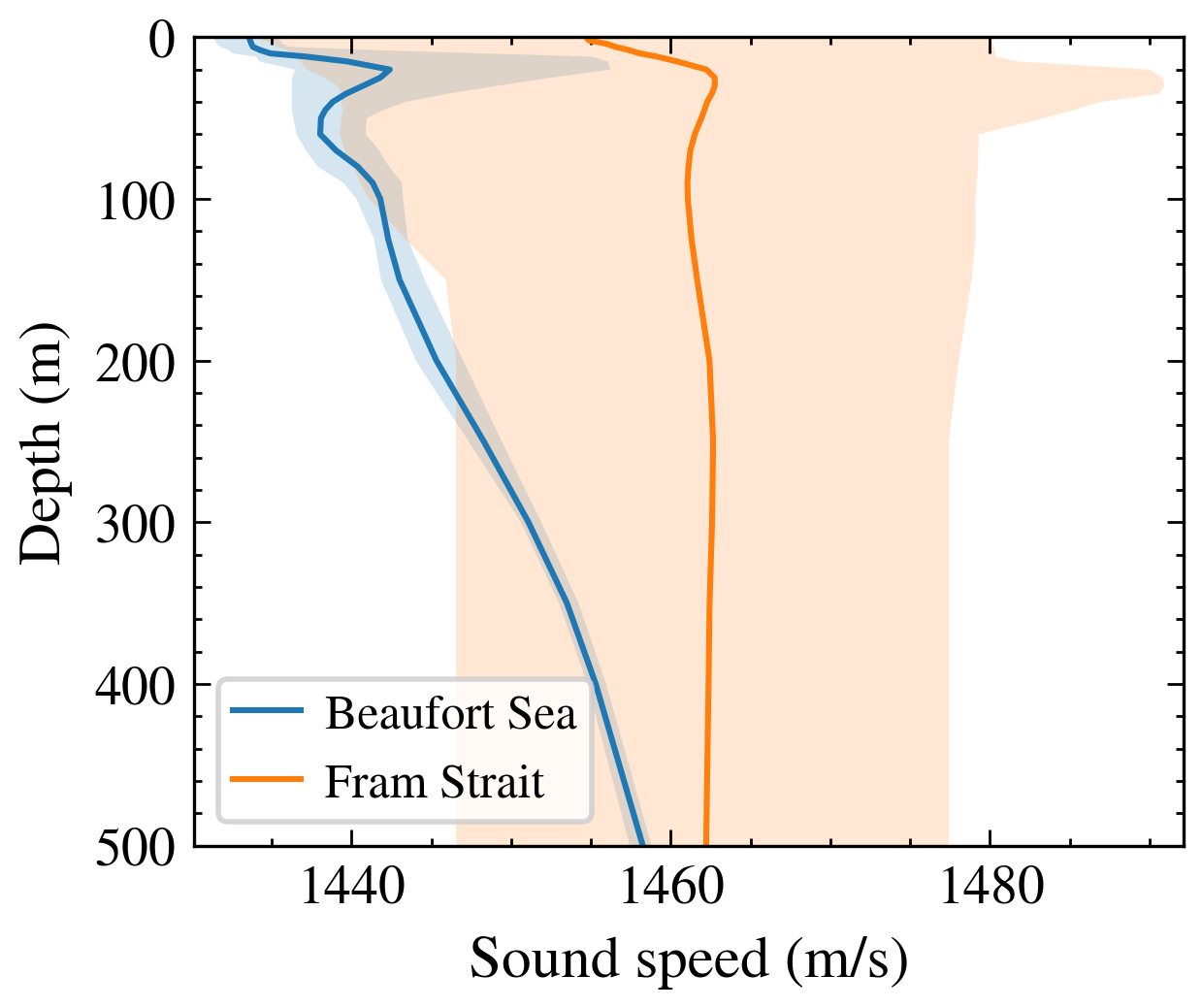}
    \caption{SSP comparison between Beaufort Sea and Fram Strait across ingested grid. Solid line is mean for each depth slice while minimum and maximum are shown by shaded region.}
    \label{fig:ssp_comparison}
    \vspace{-5mm}
\end{figure}

Two contrasting Arctic study regions, the Beaufort Sea and the Fram Strait,
were extracted from a single mid-summer snapshot (15 July 2014) of
the HYCOM reanalysis \cite{HYCOM} product to evaluate the
difference in estimators under two sound-speed regimes. Both regions share the
HYCOM native resolution of $1/12^\circ$ longitudinal $\times$ $1/25^\circ$
latitudinal in the horizontal and 40 vertical levels spanning 0 to 5000~m,
with the depth grid non-uniformly spaced. Region-specific properties are
summarized in Table~\ref{tab:study_regions}. The Beaufort Sea window centers
on the western-Arctic basin's halocline region, while the Fram
Strait window sits over the two-way passage between polar surface water and
Atlantic inflow \cite{rudels2013}, producing a substantially different sound-speed structure shown in Figure~\ref{fig:ssp_comparison} .

\begin{table}[t]
  \centering
  \caption{Physical characteristics of the two Arctic study regions extracted
from HYCOM GLBv0.08}
  \label{tab:study_regions}
  \begin{tabular}{|l|c|c|}
    \hline
    Property & Beaufort Sea & Fram Strait \\
    \hline
    Centre (lat, lon)      & $74.5^\circ$~N, $140^\circ$~W & $78^\circ$~N,
$0^\circ$~E \\
    ROI extent             & $650 \times 330$~km          & $490 \times
330$~km \\
    Grid ($x \times y$)    & $251 \times 76$              & $237 \times 75$ \\
    Regime                 & Cold-cap halocline           & Atlantic-polar
exchange \\
    \hline
  \end{tabular}
\end{table}

\begin{figure}
    \centering
    \includegraphics[width=0.85\linewidth]{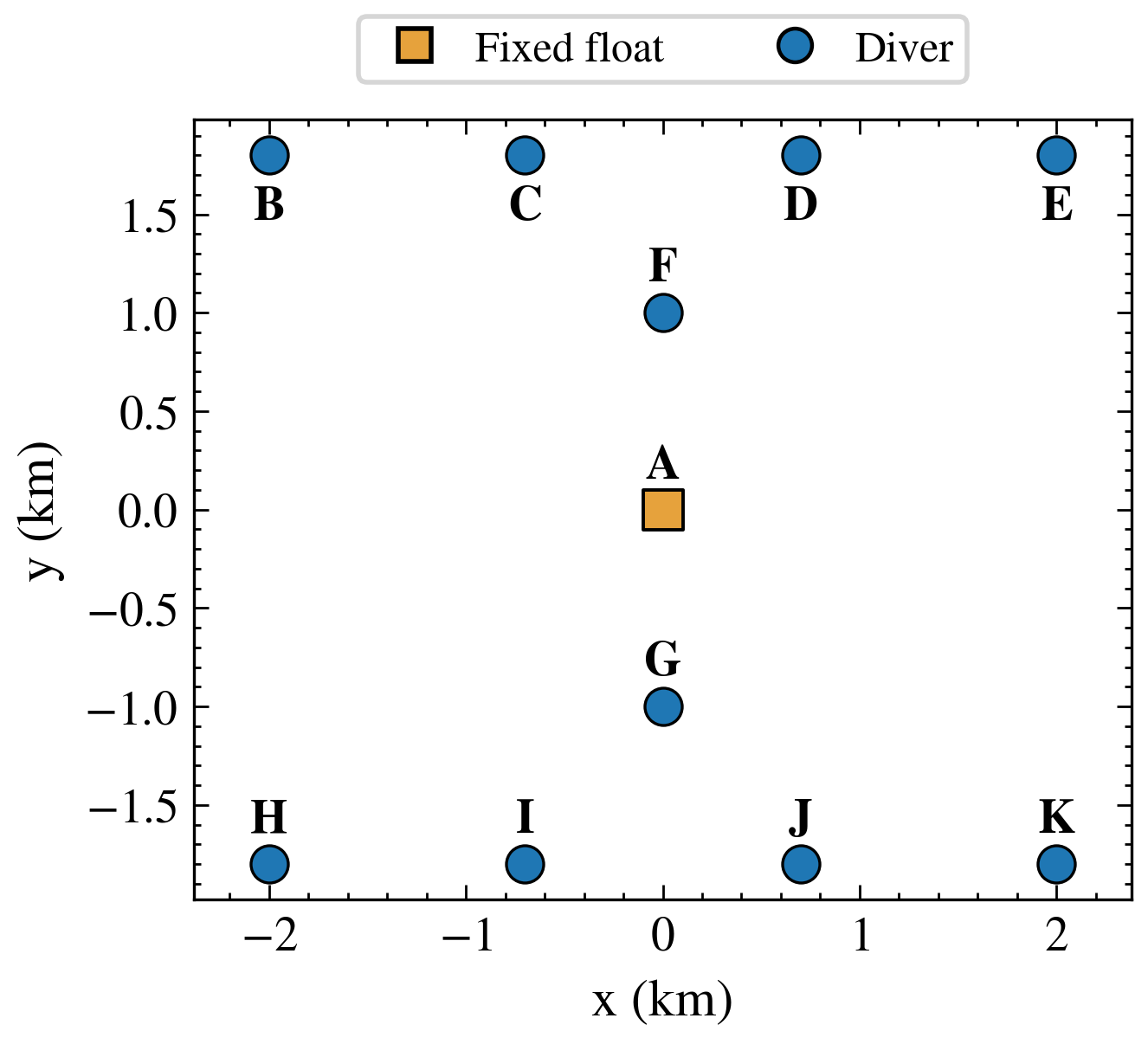}
    \caption{Fleet layout in the x-y plane with z representing depth. Float A is a station keeping agent with other current drifting floats around it in a grid.}
    \label{fig:fleet_layout}
    \vspace{-5mm}
\end{figure}

Eleven floats operate in a multi-agent fleet for the simulated experiments. Ten floats execute an Argo-style vertical cycle: they descend at a constant vertical rate to an assigned target depth, hold for a prescribed bottom dwell, ascend at the same rate, and hold briefly at the surface before repeating.
Only the vertical component of the diver's body-frame velocity is commanded; its horizontal motion is inherited from the local ocean-current field, so lateral displacement emerges passively as current drift rather than from any propulsion model.
A single station-holding beacon (agent A) is anchored at the surface serves as a fixed reference against which the floats' drifting range paths are anchored.  
The remaining 10 agents (B--K) are assigned distinct target depths spanning
250--2500~m in 250~m increments (G: 250; B: 500; D: 750; I: 1000;
K: 1250; F: 1500; J: 1750; C: 2000; E: 2250; H: 2500~m), tiling the
water column from the near-surface halocline down to the deep
pressure-dominated layer. Each float is given a initial timing offset on its
descent so that the fleet is not synchronized, enabling the effects of ranging to be studied across a variety of depths 
which induce refraction.
The resulting deployment geometry is shown in
Figure~\ref{fig:fleet_layout}. The mission duration, cycle, ping, and
sensor-sampling parameters used for both environments are defined in
Table~\ref{tab:mission_timing}.

\begin{table}[t]
  \centering
  \caption{Mission timing and sensing parameters shared by the Beaufort Sea
  and Fram Strait experiments.}
  \label{tab:mission_timing}
  \begin{tabular}{l|l}
    Parameter & Value \\
    \hline
    Mission duration                 & 168 h (7 days) \\
    Physics timestep                 & 0.125 s \\
    \hline
    Hold time at depth         & 24 h \\
    Hold time at surface       & 1 h \\
    Vertical speed             & 0.5 m/s \\
    \hline
    Sensor sampling period (Odometry, GNSS) & 200 s \\
    Acoustic ping interval           & 4 h \\
    \hline
    Monte Carlo realizations         & 25 \\
  \end{tabular}
\end{table}

  The concrete noise levels applied to each measurement stream are listed
in Table~\ref{tab:noise_params}.
The odometry rotation distribution is set two orders of magnitude tighter than
translation because current drift imparts negligible attitude change on
a passive float; the same rotation numbers apply to both floats and to
the surface beacon.

\begin{table}[t]
  \centering
  \caption{Measurement noise parameters shared by both study regions and
  every Monte Carlo realization}
  \label{tab:noise_params}
  \begin{tabular}{l|l|l}
    Measurement & Parameter & Value \\
    \hline
    Range              & $\sigma_r$                     & $1.0$ m \\
    Depth prior        & $\sigma_z$                     & $0.01$ m \\
    GNSS prior (per axis) & $\sigma_{\text{GNSS}}$      & $0.10$ m \\
    \hline
    Odom. rotation     & fraction $f_{\text{rot}}$      & $1{\times}10^{-6}$
\\
    Odom. rotation     & drift $d_{\text{rot}}$         & $5{\times}10^{-7}$
rad/s \\
    Odom. translation, xy & fraction $f_{xy}$           & $0.05$ \\
    Odom. translation, z  & fraction $f_z$              & $0.001$ \\
    Odom. translation  & drift $d_{\text{trans}}$       & $0.0125$ m/s \\
  \end{tabular}
\end{table}

\section{Results and Discussion}
\label{sec:results}

\begin{figure*}[t]
  \centering
  \includegraphics[width=0.48\linewidth]{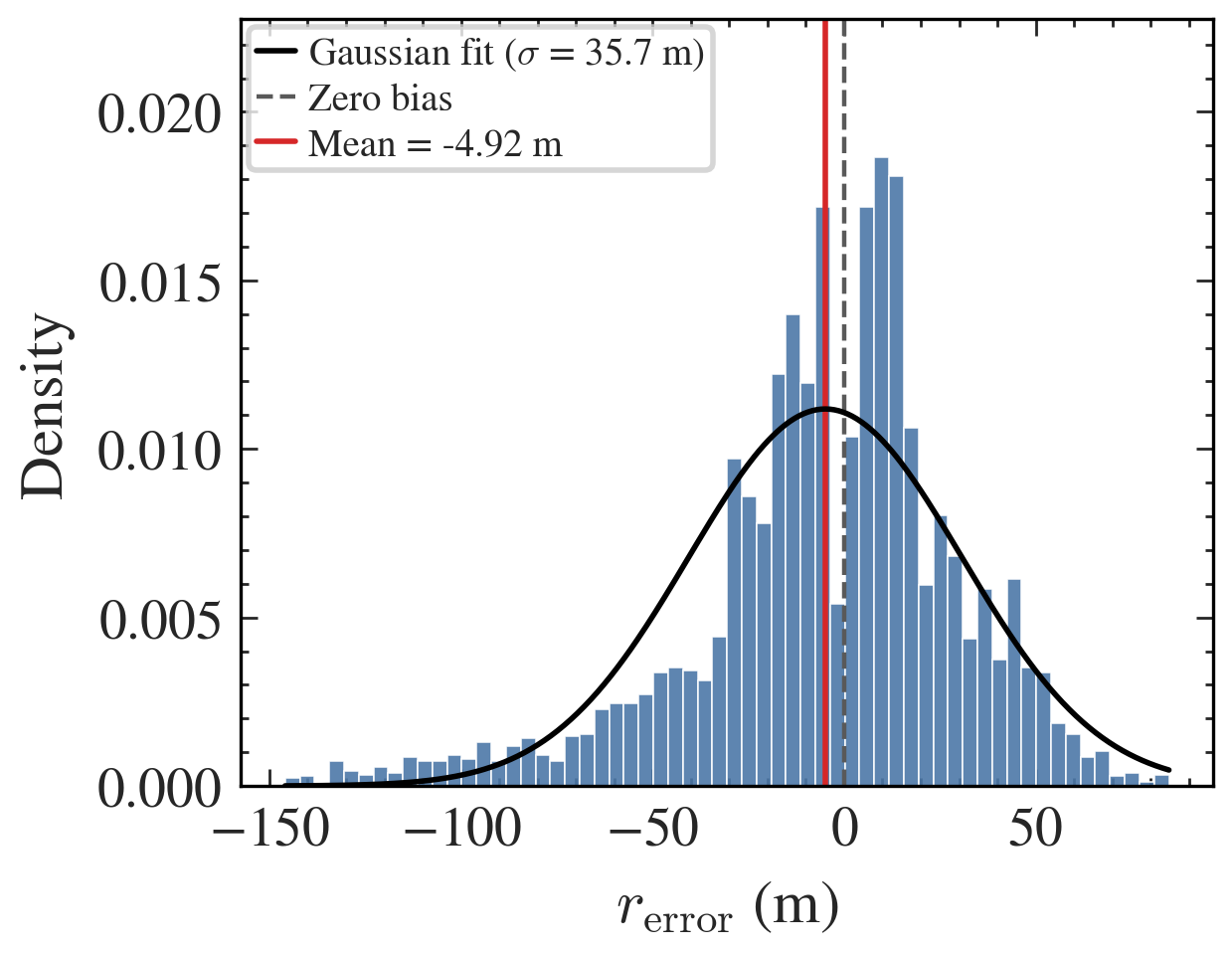}
  \includegraphics[width=0.48\linewidth]{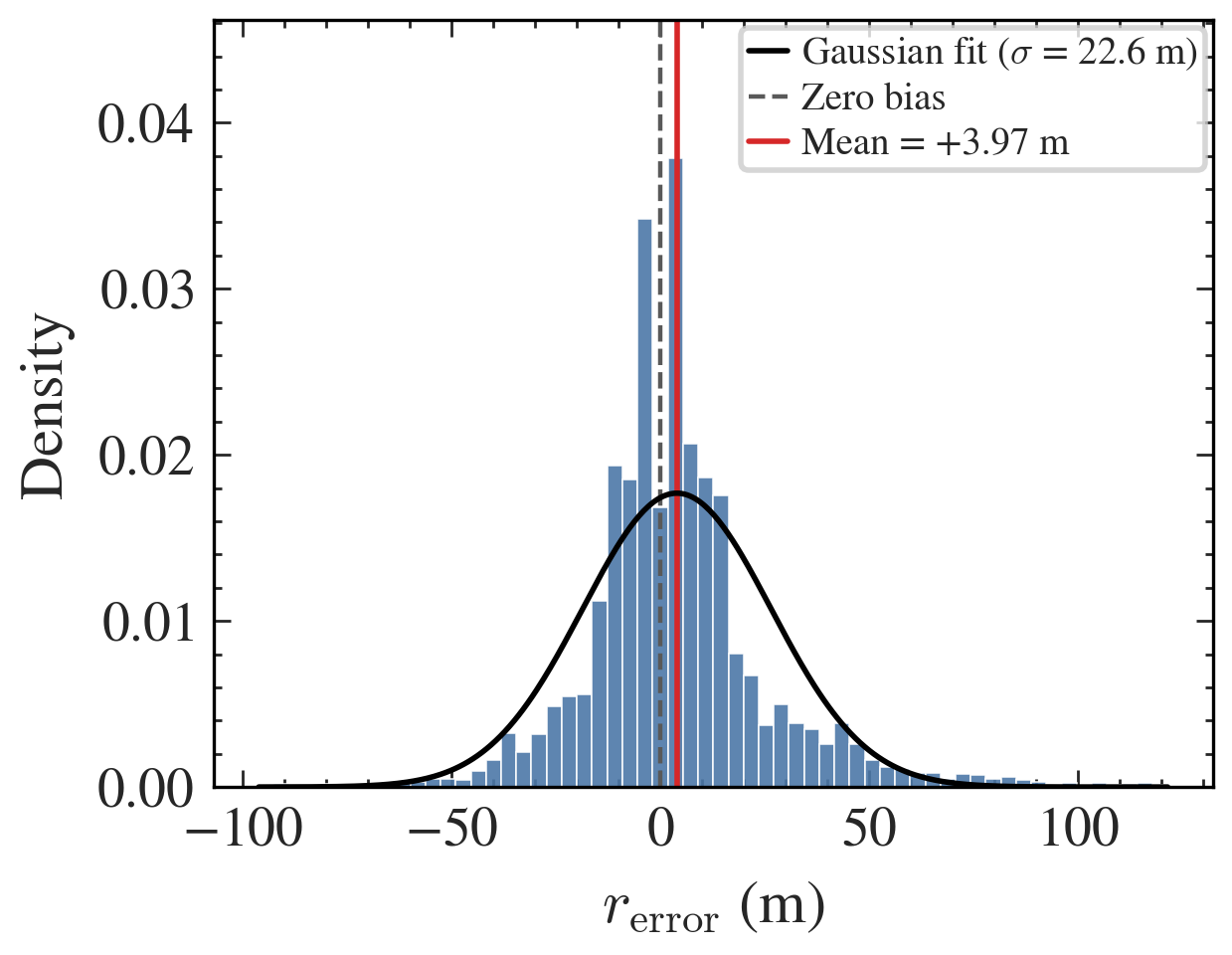}
  \caption{Fleet-wide signed range error $r_{\text{error}} =
  r_{\text{refracted}} - r_{\text{straight-line}}$ over the full
  seven-day mission. Beaufort Sea (left) with 4,574 measurements and Fram Strait (right) with 4,306 measurements.
  Solid red line is the empirical mean; dashed line marks zero. A best-fit Gaussian distribution is plotted using a solid black line.}
  \label{fig:range_bias}
  \vspace{-3mm}
\end{figure*}

\subsection{Ranging}
The empirical mean of the ranging error, $r_{\text{error}} = r_{\text{refracted}} - r_{\text{straight-line}}$, is small in both environments. The error is 
$-4.92$~m in the Beaufort Sea and $+3.97$~m in the Fram Strait, and carries
opposite signs across regions (Fig.~\ref{fig:range_bias}). However, the
distributions themselves extend to roughly $\pm 100$~m and
carry visibly heavy tails, so their empirical spread is more than
an order of magnitude larger than the $\sigma_r = 1$~m Gaussian noise
model supplied to the estimator. In both regions the ranging first moment (mean) is close to what the straight-line assumption predicts, while its second moment (variance) is under-estimated by the factor graph estimator that treats every range error as a $\mathcal{N}(0,1)$~m. Ranging errors additionally show that each oceanic environment presents a different distribution shape. The Beaufort Sea carries a heavier negative tail while the Fram Strait is more symmetric in comparison.

Table~\ref{tab:range_drop_category_compare} shows that modeling acoustic refraction reduces available ranging measurements to the estimator. The Fram Strait is particularly sensitive to ranging dropouts and overall sees greater reductions across all categories. Although this behavior is not generalizable as SSP is a time-varying oceanic property, the results demonstrate the effects acoustic shadow zones can have on available measurements for estimators. Our tool provides an environmentally informed way to identify which range readings are not available for future work in RA-SLAM.

\begin{table}[t]
\centering
\caption{Ranging failure rate (\%) by link category. (4{,}620 total ranging attempts
each)}
\label{tab:range_drop_category_compare}
\begin{tabular}{l|r|r}
  Link category & Beaufort & Fram Strait \\
  \hline
  Station-keeping Float $\leftrightarrow$ Float & 3.6 & 29.5 \\
  Float $\leftrightarrow$ Float & 0.4 & 1.7 \\
  \hline
  All links & 1.0 & 6.8 \\
\end{tabular}
\end{table}

\subsection{Trajectory error}
\begin{figure}[t]
  \centering
  \includegraphics[width=0.49\linewidth]{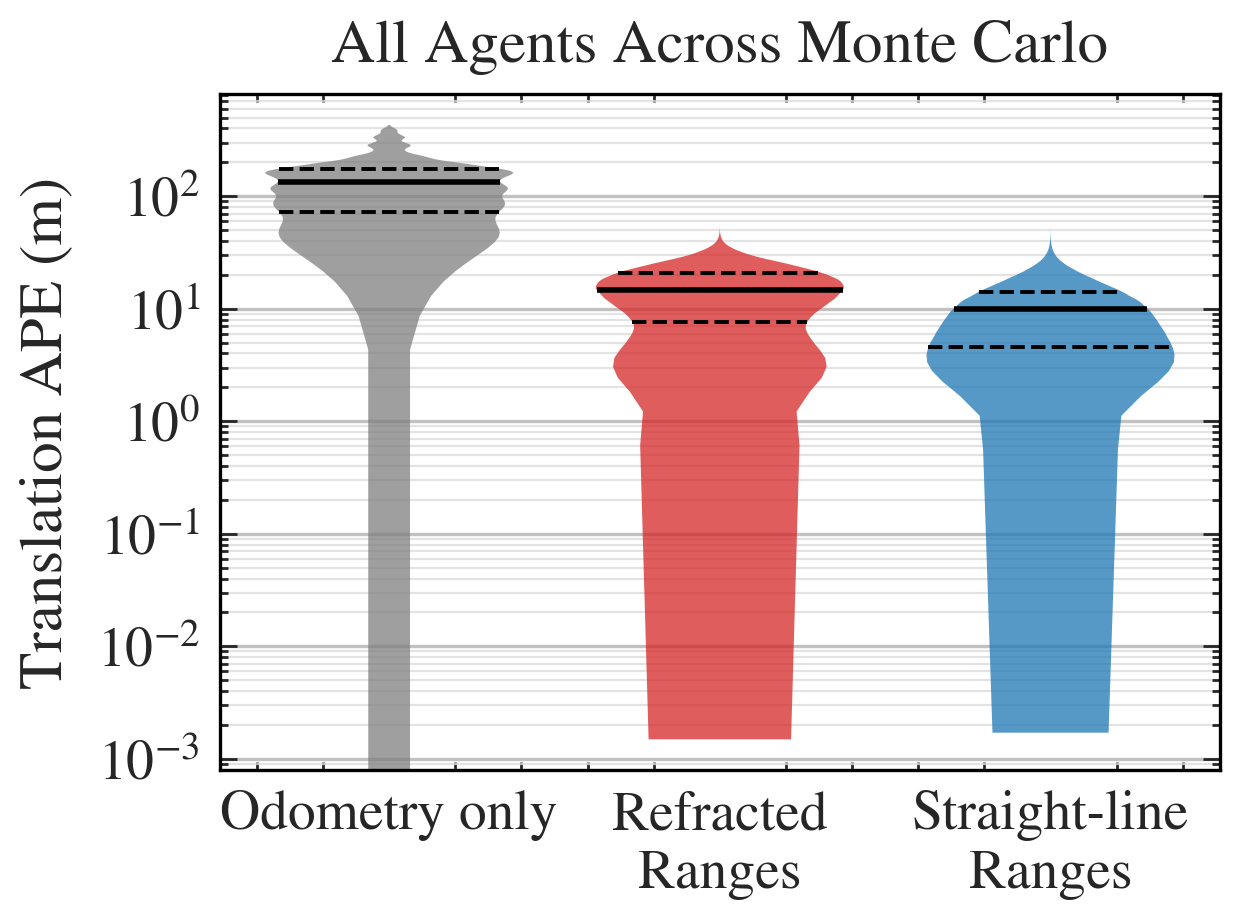}
  \includegraphics[width=0.49\linewidth]{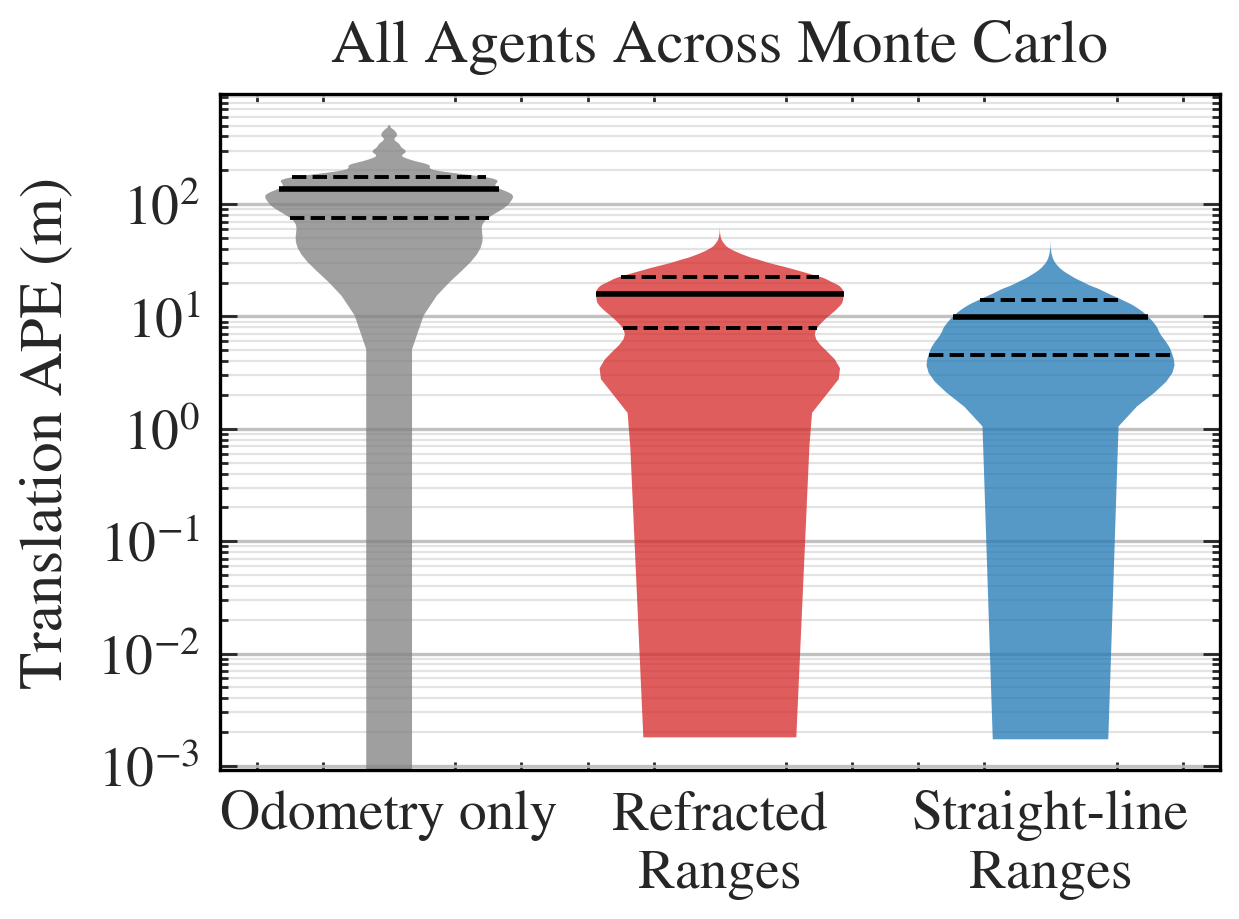}
  \caption{Aggregate translation APE across all divers and Monte Carlo
  seeds under three ranging conditions. Beaufort Sea (left), Fram
  Strait (right). Solid line marks the mean, dashed lines mark the
  25th/75th percentiles.}
  \label{fig:pooled_ape}
  \vspace{-3mm}
\end{figure}

\begin{figure}[t]
  \centering
  \includegraphics[width=0.80\linewidth]{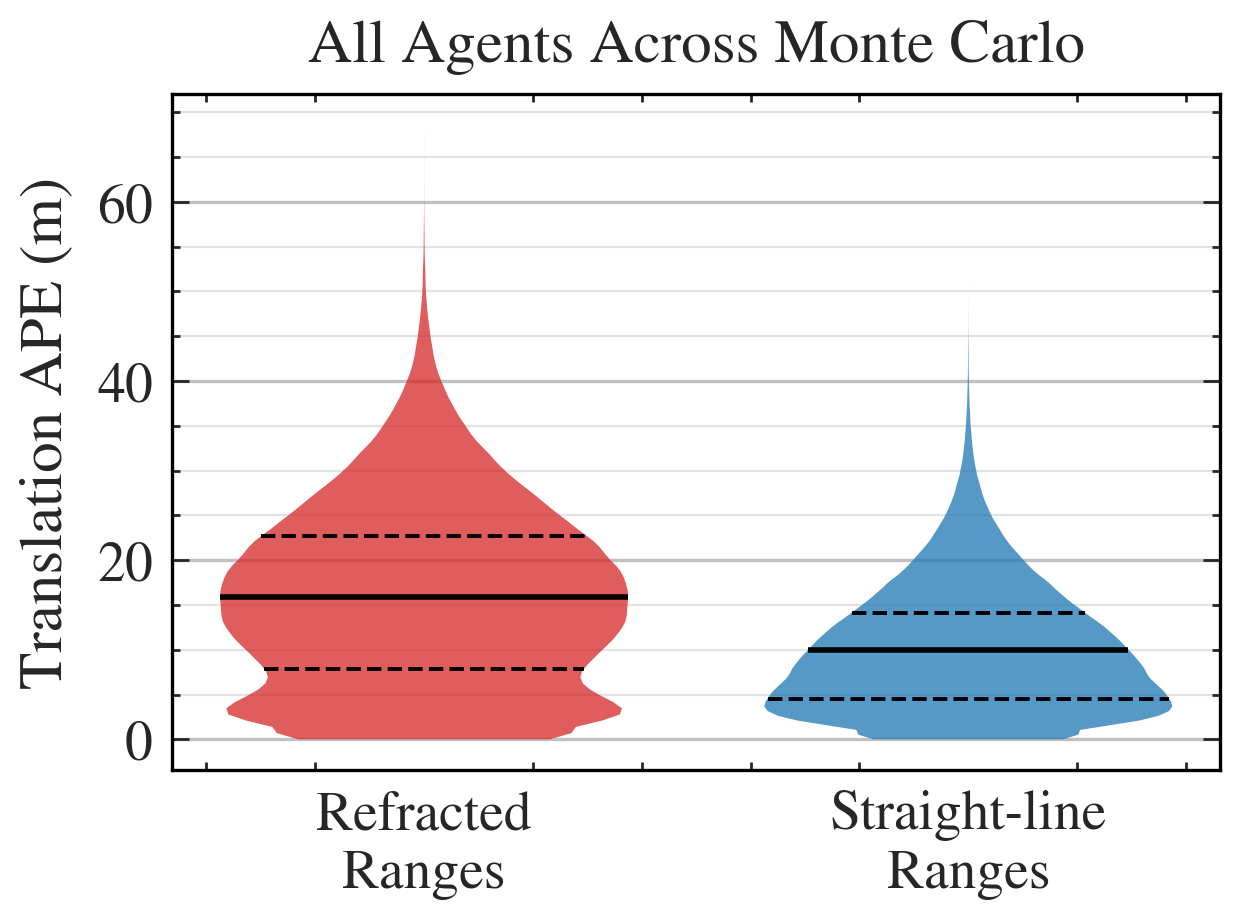}
  \caption{Aggregate translation APE in Fram Strait (non-log scale)}
  \label{fig:pooled_ape_fram}
  \vspace{-3mm}
\end{figure}

Aggregate ATE across every agent and
every Monte Carlo case (Fig.~\ref{fig:pooled_ape}) shows that ATE distributions for straight-line ranges have lower mean values and shorter tails than refracted ranges. Both measurement types when fused in the estimator outperform naive odometry. The refracted range ATE distribution's overall shape differences (Fig.~\ref{fig:pooled_ape_fram}), mean, and percentile values show how refraction systematically biases the estimator when compared to straight-line measurements given the same estimator. A likely cause of systematic bias in the estimator is the heavy tail of $r_{\text{error}}$ which has an outsized influence on the estimator through the larger residuals associated with those measurements.

Simulated experiments evaluated over the Beaufort Sea and the Fram Strait show that as collaborative fleets scale to operational domains that span 5-12 kilometers, acoustic refraction will begin to systematically bias estimators. The floats presented in this work were simplifications intended to exercise the SSP profile in the region; however, more complex trajectories may further change results and will be a subject of future work. Our work also presents a foundation for which future environmentally-informed estimators can be evaluated against to determine if they are capable of removing the systematic bias caused by refraction.

\section{Conclusion}
This works presents a set of simulated experiments in the Beaufort Sea and the Fram Strait that demonstrate refracted range measurements bias naive multi-agent factor graph estimators. The key component that enabled achieving this goal was the development of \texttt{MantaRay}, a simulator capable of leveraging both well-established Bellhop acoustic simulation methods and HYCOM reanalysis data to deliver environmentally-informed acoustic ranging measurements for collaborative fleets of agents. This methodology additionally demonstrated that including environmentally-informed refraction captures acoustic shadow zones and reduces the number of measurements available to estimators compounding the challenge. Although environmental results are dependent on the specific SSP in the time period sampled, the presented results stand as grounds for further investigation in operating regimes and time periods of interest.

\textbf{Acknowledgments:} The authors would like to thank John Folkesson for mentorship and support during this work.

\printbibliography

\end{document}